\documentclass{article}

\usepackage[nonatbib,final]{neurips_2020}

\usepackage[utf8]{inputenc} 
\usepackage[T1]{fontenc}    
\usepackage{hyperref}       
\usepackage{url}            
\usepackage{booktabs}       
\usepackage{amsfonts}       
\usepackage{amssymb}
\usepackage{amsmath}
\usepackage{nicefrac}       
\usepackage{microtype}      
\usepackage{multirow}       
\usepackage{makecell}       
\usepackage[dvipsnames]{xcolor}
\usepackage{pifont}
\usepackage{listings}
\usepackage{wrapfig}
\usepackage{multicol}
\usepackage{float}
\usepackage{paracol} 
\usepackage{todonotes}
\usepackage[ruled,vlined]{algorithm2e}
\usepackage{algpseudocode}
\usepackage{capt-of}
\usepackage{subcaption}
\usepackage{enumitem}
\usepackage{placeins}

\def\NAME{{PyGlove}}
\def\Figref#1{Figure~\ref{#1}}

\def\Secref#1{Section~\ref{#1}}

\def\Tabref#1{Table~\ref{#1}}
\def\sspace{{\mathcal{S}}}
\def\nasone{{NAS-Bench-101}}
\newcommand{\code}[1]{\texttt{#1}}

\usepackage{cite}

\newif\ifhidetodos \hidetodosfalse  

\definecolor{darkgreen}{rgb}{0,0.6,0}

\definecolor{lightyellow}{HTML}{FFFFE0}
\newcommand{\highlight}[1]{\hspace{-0.9mm}\colorbox{lightyellow}{#1}}
\definecolor{darkgreen}{HTML}{006400}

\lstdefinestyle{Python}{
    basicstyle      = \ttfamily \lst@ifdisplaystyle\footnotesize\fi,
    keywordstyle    = \color{black},
    stringstyle     = \color{black},
    commentstyle    = \color{gray}\ttfamily,
    frame=single,
    framerule=0.0pt,
    numbers=none,
    numbersep=0.2cm
}
\title{PyGlove: Symbolic Programming\\for Automated Machine Learning}

\author{
Daiyi Peng, Xuanyi Dong, Esteban Real,
Mingxing Tan, Yifeng Lu\\
\textbf{Hanxiao Liu, Gabriel Bender, Adam Kraft, Chen Liang, Quoc V. Le}\\
\vspace{0.3em}
Google Research, Brain Team\\
\small{\{daiyip, ereal, tanmingxing, yifenglu,} \\ \small{hanxiaol, gbender, adamkraft, crazydonkey, qvl\}@google.com} \\
\vspace{0.5em}
\small{xuanyi.dxy@gmail.com \thanks{Work done as a research intern at Google.}}
}

\begin{document}

\maketitle

\newcommand{\revision}{final}

\vspace{-15pt}
\begin{abstract}
Neural networks are sensitive to hyper-parameter and architecture choices. Automated Machine Learning (AutoML) is a  promising paradigm for automating these choices.
Current ML software libraries, however, are quite limited in handling the dynamic interactions among the components of AutoML.
For example, efficient NAS algorithms,
such as ENAS~\cite{pham2018efficient} and DARTS~\cite{liu2018darts}, typically require an implementation coupling between the search space and search algorithm, the two key components in AutoML. Furthermore, implementing a complex search flow, such as searching architectures within a loop of searching hardware configurations, is difficult.
To summarize, changing the search space, search algorithm, or search flow in current ML libraries usually requires a significant change in the program logic.\looseness=-1

In this paper, we introduce a new way of programming AutoML based on symbolic programming. Under this paradigm, ML programs are mutable, thus can be manipulated easily by another program. As a result, AutoML can be reformulated as an automated process of symbolic manipulation. With this formulation, we decouple the triangle of the search algorithm, the search space and the child program. This decoupling makes it easy to change the search space and search algorithm (without and with weight sharing), as well as to add search capabilities to existing code and implement complex search flows. We then introduce PyGlove, a new Python library that implements this paradigm. Through case studies on ImageNet and {\nasone}, we show that with PyGlove users can easily convert a static program into a search space, quickly iterate on the search spaces and search algorithms, and craft complex search flows to achieve better results.

\end{abstract}
\vspace{-5pt}
\section{Introduction}
Neural networks are sensitive to architecture and hyper-parameter choices~\cite{melis2018state,canziani2016analysis}. For example, on the ImageNet dataset \cite{krizhevsky2012imagenet}, we have observed a large increase in accuracy thanks to changes in architectures, hyper-parameters, and training algorithms, from the seminal work of AlexNet~\cite{krizhevsky2012imagenet} to  recent state-of-the-art models such as EfficientNet~\cite{tan2019efficientnet}. However, as neural networks become increasingly complex, the potential number of architecture and hyper-parameter choices becomes numerous. Hand-crafting neural network architectures and selecting the right hyper-parameters is, therefore,  increasingly difficult and often take months of experimentation.

Automated Machine Learning (AutoML) is a promising paradigm for tackling this difficulty. In AutoML, selecting architectures and hyper-parameters is formulated as a search problem, where a \emph{search space} is defined to represent all possible choices and a \emph{search algorithm} is used to find the best choices. For hyper-parameter search, the search space would specify the range of values to try. For architecture search, the search space would specify the architectural configurations to try. The search space plays a critical role in the success of neural architecture search (NAS)~\cite{zoph2018learning,tan2019mnasnet}, and can be significantly different from one application to another~\cite{zoph2017neural,tan2019mnasnet,wu2019fbnet, dai2019chamnet}. In addition, there are also many different search algorithms, such as random search~\cite{bergstra2012random}, Bayesian optimization~\cite{snoek2012practical}, RL-based methods~\cite{zoph2017neural, pham2018efficient, cai2018proxylessnas, bender2020tunas}, evolutionary methods~\cite{real2019regularized}, gradient-based methods~\cite{liu2018darts, xie2020snas, wu2019fbnet} and neural predictors~\cite{wen2019neural}.\looseness=-1

This proliferation of search spaces and search algorithms in AutoML makes it difficult to program with existing software libraries. In particular, a common problem of current libraries is that search spaces and search algorithms are tightly coupled, making it hard to modify search space or search algorithm alone. A practical scenario that arises is the need to upgrade a search algorithm while keeping the rest of the infrastructure the same. For example, recent years have seen a transition from AutoML algorithms that train each model from scratch~\cite{zoph2017neural,tan2019mnasnet} to those that employ weight-sharing to attain massive efficiency gains, such as ENAS and DARTS \cite{pham2018efficient,liu2018darts,cai2018proxylessnas,dong2020autohas,bender2020tunas}. Yet, upgrading an existing search space by introducing weight-sharing requires significant changes to both the search algorithm and the model building logic, as we will see in \Secref{subsec:disentangle}. Such coupling between search spaces and search algorithms, and the resulting inflexibility, impose a heavy burden on AutoML researchers and practitioners.\looseness=-1

\vspace{-2pt}

We believe that the main challenge lies in the programming paradigm mismatch between existing software libraries and AutoML. Most existing libraries are built on the premise of immutable programs, where a fixed program is used to process different data. On the contrary, AutoML requires programs (i.e. model architectures) to be mutable, as they must be dynamically modified by another program (i.e. the search algorithm) whose job is to explore the search space. Due to this mismatch, predefined interfaces for search spaces and search algorithms struggle to accommodate unanticipated interactions, making it difficult to try new AutoML approaches. \emph{Symbolic programming}, which originated from LISP~\cite{mccarthy1960recursive}, provides a potential solution to this problem, by allowing a program to manipulate its own components as if they were plain data~\cite{spwiki}. However, despite its long history, symbolic programming has not yet been widely explored in the ML community.

\vspace{-2pt}

In this paper, we reformulate AutoML as an automated process of manipulating ML programs symbolically. Under this formulation, programs are mutable objects which can be cloned and modified after their creation. These mutable objects can express standard machine learning concepts, from a convolutional unit to a complex user-defined training procedure. As a result, all parts of a ML program are mutable. Moreover, through symbolic programming, programs can modify programs. Therefore the interactions between the child program, search space, and search algorithm are no longer static. We can mediate them or change them via meta-programs. For example, we can map the search space into an abstract view which is understood by the search algorithm, translating an architectural search space into a super-network that can be optimized by efficient NAS algorithms.

\vspace{-2pt}

Further, we propose \emph{PyGlove}, a library that enables general symbolic programming in Python, as an implementation of our method tested on real-world AutoML scenarios. With PyGlove, Python classes and functions can be made mutable through brief Python annotations, which makes it much easier to write AutoML programs. PyGlove allows AutoML techniques to be easily dropped into preexisting ML pipelines, while also benefiting open-ended research which requires extreme flexibility.

\vspace{-2pt}

To summarize, our contributions are the following:
\vspace{-5pt}
\begin{itemize}
\itemsep0em
  \item We reformulate AutoML under the symbolic programming paradigm, greatly simplifying the programming interface for AutoML by accommodating unanticipated interactions among the child programs, search spaces and search algorithms via a mutable object model.
  \item We introduce \emph{PyGlove}, a general symbolic programming library for Python which implements our symbolic formulation of AutoML. With PyGlove, AutoML can be easily dropped into preexisting ML programs, with all program parts searchable, permitting rapid exploration on different dimensions of AutoML.
  \item Through case studies, we demonstrate the expressiveness of PyGlove in real-world search spaces. We demonstrate how PyGlove allows AutoML researchers and practitioners to change search spaces, search algorithms and search flows with only a few lines of code.
\end{itemize}

\vspace{-7pt}
\section{Symbolic Programming for AutoML}
\label{sec:method}

\vspace{-5pt}

Many AutoML approaches (e.g., \cite{zoph2017neural,real2017large,liu2018darts}) can be formulated as three interacting components: the \emph{child program}, the \emph{search space}, and the \emph{search algorithm}. AutoML's goal is to discover a performant child program (e.g., a neural network architecture or a data augmentation policy) out of a large set of possibilities defined by the search space. The search algorithm accomplishes the said goal by iteratively sampling child programs from the search space. Each sampled child program is then evaluated, resulting in a numeric measure of its quality. This measure is called the \emph{reward}\footnote{While we use RL concepts to illustrate the core idea of our method, as will be shown later, the proposed paradigm is applicable to other types of AutoML methods as well.}. The reward is then fed back to the search algorithm to improve future sampling of child programs.

In typical AutoML libraries~\cite{NIPS2015_5872,kotthoff2017auto,feurer2019auto,golovin2017google,NNI,agtabular,negrinho2019towards,keras_tuner,jin2019auto}, these three components are usually tightly coupled. The coupling between these components means that we cannot change the interactions between them unless non-trivial modifications are made.   
This limits the flexibility of the libraries. Some successful attempts have been made to break these couplings. For example, Vizier~\cite{golovin2017google} decouples the search space and the search algorithm by using a dictionary as the search space contract between the child program and the search algorithm, resulting in modular black-box search algorithms. Another example is the NNI library~\cite{NNI}, which tries to unify search algorithms with and without weight sharing by carefully designed APIs. This paper, however, solves the coupling problem in a different and more general way: with symbolic programming, programs are allowed to be modified by other programs. Therefore, instead of solving fixed couplings, we allow dynamic couplings through a mutable object model. In this section, we will explain our method and show how this makes AutoML programming more flexible.\looseness=-1

\vspace{-5pt}
\subsection{AutoML as an Automated Symbolic Manipulation Process}
\label{subsec:symbolic_process}
\vspace{-5pt}

\begin{figure}[H]
    \vspace{-5pt}
    \centering
    \includegraphics[width=\linewidth]{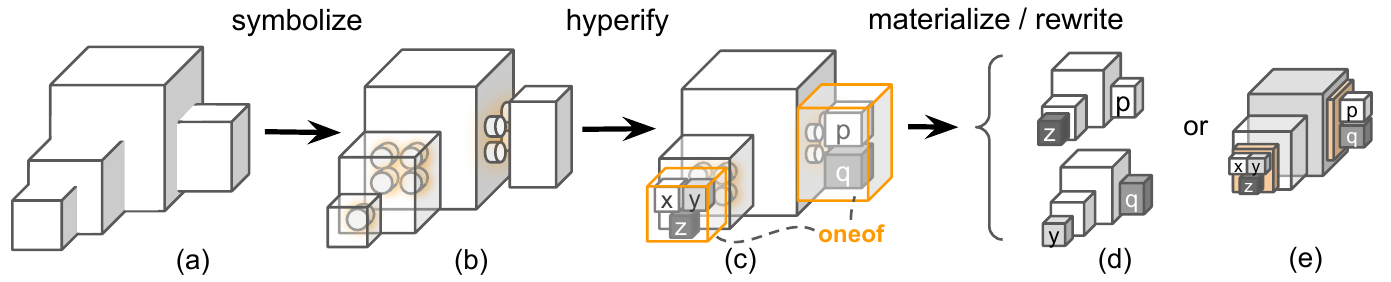}
    \vspace{-15pt}
    \caption{\footnotesize{
    AutoML as an automated symbolic manipulation process.
    }}
    \label{fig:overview}
    \vspace{-15pt}
\end{figure}

AutoML can be interpreted as an automated process of searching for a child program from a search space to maximize a reward. We decompose this process into a sequence of symbolic operations. A (regular) child program (\Figref{fig:overview}-a) is \emph{symbolized} into a symbolic child program (\Figref{fig:overview}-b), which can be then cloned and modified. The symbolic program is further \emph{hyperified} into a search space (\Figref{fig:overview}-c) by replacing some of the fixed parts with to-be-determined specifications. During the search, the search space is \emph{materialized} into different child programs (\Figref{fig:overview}-d) based on search algorithm decisions, or can be rewritten into a super-program (\Figref{fig:overview}-e) to apply complex search algorithms such as efficient NAS. 

An analogy to this process is to have a robot build a house with LEGO~\cite{lego} bricks to meet a human being's taste: symbolizing a regular program is like converting molded plastic parts into LEGO bricks; hyperifying a symbolic program into a search space is like providing a blueprint of the house with variations. With the help of the search algorithm, the search space is materialized into different child programs whose rewards are fed back to the search algorithm to improve future sampling, like a robot trying different ways to build the house and gradually learning what humans prefer.

\begin{wrapfigure}{r}{0.46\linewidth}
    \vspace{-13pt}
    \centering
    \includegraphics[width=\linewidth]{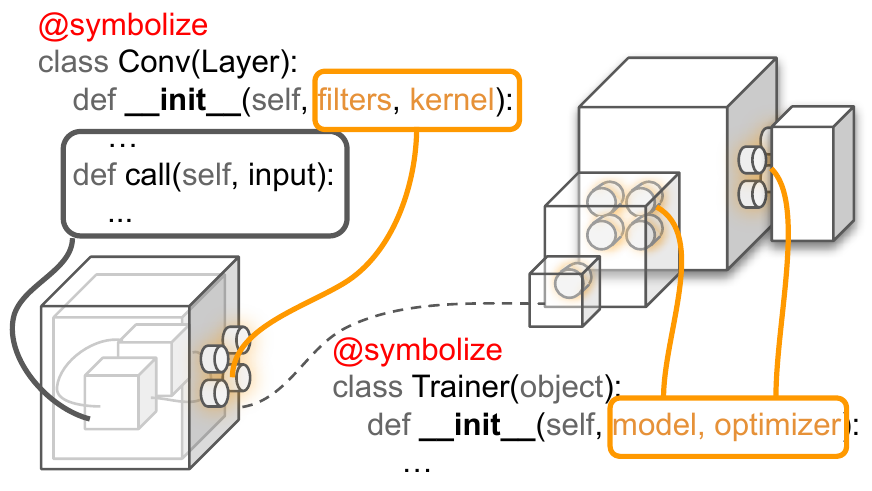}
    \caption{
    \footnotesize{
    Symbolizing classes into mutable symbolic trees. Their hyper-parameters are like the studs of LEGO bricks, while their implementations are less interesting while we manipulate the trees.\looseness=-1
    }}
    \label{fig:symbolize}
    \vspace{-10pt}
\end{wrapfigure}

\paragraph{Symbolization.} A (regular) child program can be described as a complex object, which is a composition of its sub-objects. A symbolic child program is such a composition whose sub-objects are no longer tied together forever, but are detachable from each other hence can be replaced by other sub-objects. The symbolic object can be hierarchical, forming a \emph{symbolic tree} which can be \emph{manipulated} or \emph{executed}. A symbolic object is manipulated through its hyper-parameters, which are like the studs of a LEGO brick, interfacing connections with other bricks. However, symbolic objects, unlike LEGO bricks, can have internal states which are automatically recomputed upon modifications. For example, when we change the dataset of a trainer, the train steps will be recomputed from the number of examples in the dataset if the training is based on the number of epochs. With such a mutable object model, we no longer need to create objects from scratch repeatedly, or modify the producers up-stream, but can clone existing objects and modify them into new ones.  The symbolic tree representation puts an emphasis on manipulating the object definitions, while leaving the implementation details behind.  \Figref{fig:symbolize} illustrates the symbolization process.\looseness=-1

\subsection{Disentangling AutoML through Symbolic Programming}
\label{subsec:disentangle}

\begin{wrapfigure}{r}{0.38\linewidth}
    \vspace{-15pt}
    \centering
    \includegraphics[width=\linewidth]{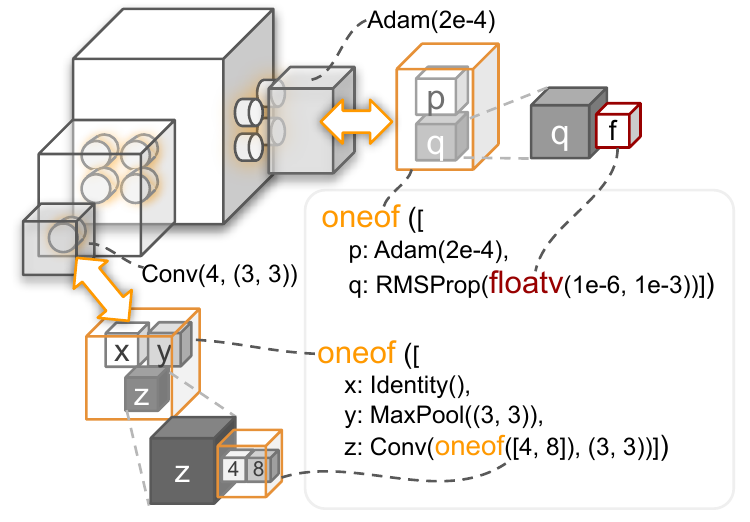}
    \caption{\footnotesize{
    Hyperifying a child program into a search space by replacing fixed parts with to-be-determined specifications.\looseness=-1
    }}
    \label{fig:hyperify}
    \vspace{5pt}
    \includegraphics[width=\linewidth]{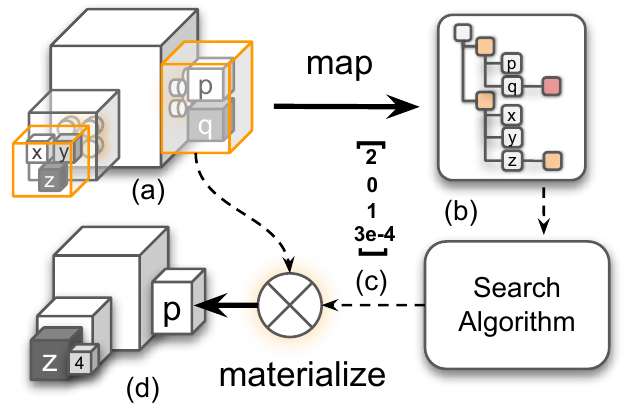}
    \vspace{-15pt}
    \caption{\footnotesize{Materializing a (concrete) child program (d) from the search space (a) with an abstract child program (c) proposed from the search algorithm, which holds an abstract search space (b) as the algorithm's view for the (concrete) search space.}
    }
    \label{fig:materialize}
    \vspace{-25pt}
\end{wrapfigure}

\paragraph{Disentangling search spaces from child programs.} The search space can be disentangled from the child program in that 1) the classes and functions of the child program can be implemented without depending on any AutoML library (Appendix B.1.1), which applies to most preexisting ML projects whose programs were started without taking AutoML in mind;  2) a child program can be manipulated into a search space without modifying its implementation. \Figref{fig:hyperify} shows that a child program is turned into a search space by replacing a \emph{fixed} \code{Conv} with a \emph{choice} of \code{Identity}, \code{MaxPool} and \code{Conv} with searchable filter size. Meanwhile, it swaps a \emph{fixed} \code{Adam} optimizer with a \emph{choice} between the \code{Adam} and an \code{RMSProp} with a searchable learning rate.

\begin{figure}[H]
    \centering
    \includegraphics[width=\linewidth]{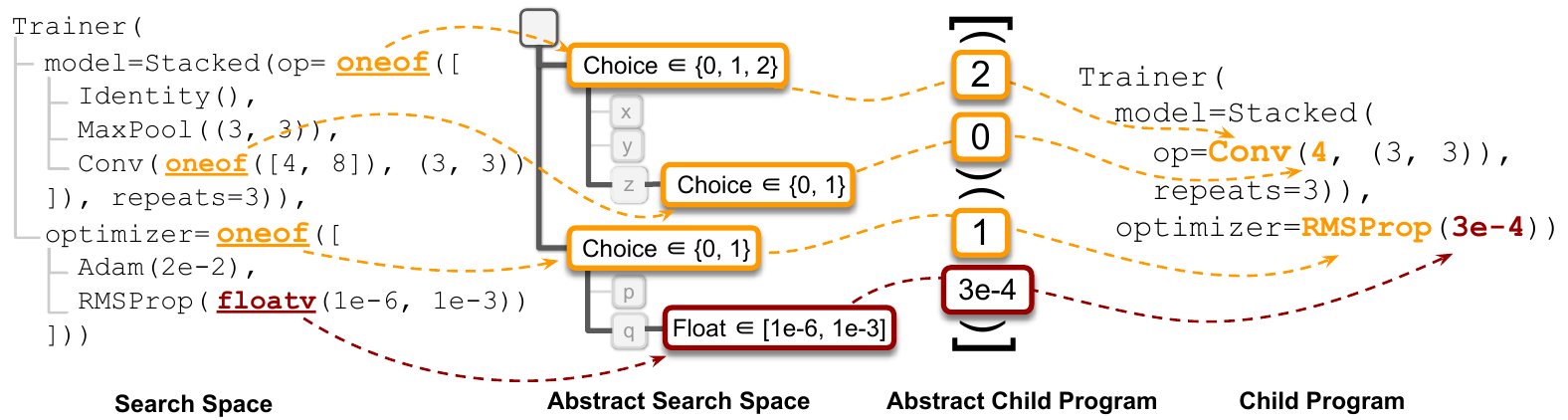}
    \caption{
    \footnotesize{
    The path from a (concrete) search space to a (concrete) child program. The disentanglement between the search space and the search algorithm is achieved by (1) abstracting the search space, (2) proposing an abstract child program, and (3) materializing the abstract child program into a concrete one.\looseness=-1
    }}
    \label{fig:materialize_details}
\vspace{-10pt}
\end{figure}

\paragraph{Disentangling search spaces from search algorithms.} Symbolic programming breaks the coupling between the search space and the search algorithm by preventing the algorithm from seeing the full search space specification. Instead, the algorithm only sees what it needs to see for the purposes of searching. We refer to the algorithm's view of the search space as the \emph{abstract search space}. The full specification, in contrast, will be called the \emph{concrete search space} (or just the ``search space'' outside this section). The distinction between the concrete and abstract search space is illustrated in \Figref{fig:materialize}: the concrete search space acts as a boilerplate for producing concrete child programs, which holds all the program details (e.g., the fixed parts). However, the abstract search space only sees the parts that need decisions, along with their numeric ranges. Based on the abstract search space, an \emph{abstract child program} is proposed, which can be static numeric values or variables. The static form is for obtaining a concrete child program, shown in \Figref{fig:materialize}, while the variable form is used for making a super-program used in efficient NAS -- the variables can be either discrete for RL-based use cases or real-valued vectors for gradient-based methods.
Mediated by the abstract search space and the abstract child program, the search algorithm can be thoroughly decoupled from the child program. \Figref{fig:materialize_details} gives a more detailed illustration of \Figref{fig:materialize}.

\vspace{-5pt}

\begin{wrapfigure}{r}{0.4\linewidth}
    \centering
    \includegraphics[width=\linewidth]{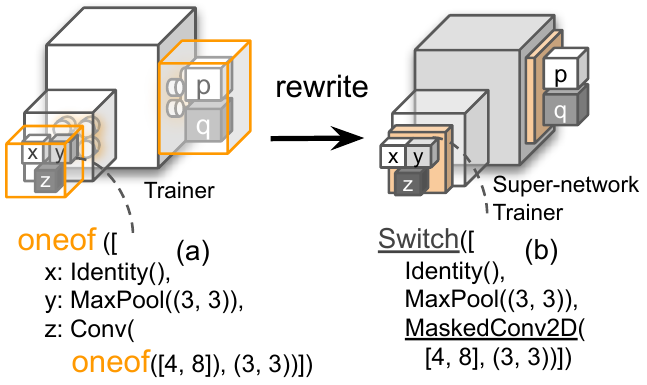}
    \vspace{-15pt}
    \caption{\footnotesize{
    Rewriting a search space (a) into a super-program (b) required by TuNAS.}
    }
    \label{fig:rewrite}
    \vspace{-10pt}
\end{wrapfigure}

\paragraph{Disentangling search algorithms from child programs.} While many search algorithms can be implemented by rewriting symbolic objects, complex algorithms such as ENAS~\cite{pham2018efficient}, DARTS~\cite{liu2018darts} and TuNAS~\cite{bender2020tunas} can be decomposed into 1) a child-program-agnostic algorithm, plus 2) a meta-program (e.g. a Python function) which rewrites the search space into a representation required by the search algorithm. The meta-program only manipulates the symbols which are interesting to the search algorithm and ignores the rest. In this way, we can decouple the search algorithm from the child program.

For example, the TuNAS~\cite{bender2020tunas} algorithm can be decomposed into 1) an implementation of REINFORCE~\cite{williams1992simple} and 2) a rewrite function which transforms the architecture search space into a super-network, and replaces the regular trainer with a trainer that samples and trains the super-network, illustrated in \Figref{fig:rewrite}. If we want to switch the search algorithm to DARTS~\cite{liu2018darts}, we use a different rewrite function that generates a super-network with soft choices, and replace the trainer with a super-network trainer that updates the choice weights based on the gradients.

\vspace{-5pt}
\subsection{Search space partitioning and complex search flows}
\label{subsec:parition_search_space}

Early work~\cite{bratman2012strong, arber2018, dong2020autohas} shows that factorized search can help partition the computation for optimizing different parts of the program. 
Yet, complex search flows have been less explored, possibly due in part to their implementation complexity. The effort involved in partitioning a search space and coordinating the search algorithms is usually non-trivial. However, the symbolic tree representation makes search space partitioning a much easier task: with a partition function, we can divide those to-be-determined parts into different groups and optimize each group separately. As a result, each optimization process sees only a portion of the search space -- a sub-space -- and they work together to optimize the complete search space.  Section~\ref{subsec:pyglove_search} discusses common patterns of such collaboration and how we express complex search flows.

\vspace{-5pt}
\section{AutoML with PyGlove}\label{sec:impl}

In this section, we introduce PyGlove, a general symbolic programming library on Python, which also implements our method for AutoML. With examples, we demonstrate how a regular program is made symbolically programmable, then turned into search spaces, searched with different search algorithms and flows in a dozen lines of code.

\begin{wraptable}{r}{0.50\textwidth}
\vspace{-36pt}
\setlength{\tabcolsep}{2pt}
\caption{\footnotesize{The development cost of dropping PyGlove into existing projects on different ML frameworks. The source code of MNIST is included in Appendix B.5.
}}
\vspace{-5pt}
\footnotesize{
\begin{tabular}{c c c}
    Projects    & \makecell{Original\\lines of code} & \makecell{Modified\\lines of code} \\
    \toprule
    PyTorch ResNet~\cite{paszke2019pytorch} & 353 & 15 \\
    \midrule
    TensorFlow MNIST~\cite{abadi2016tensorflow} & 120 & 24\\
    \bottomrule
\end{tabular}
}
\vspace{-10pt}
\label{table:drop_in_cost}
\end{wraptable}
\vspace{-5pt}
\begin{figure}[h!]
\begin{center}
\includegraphics[width=\linewidth]{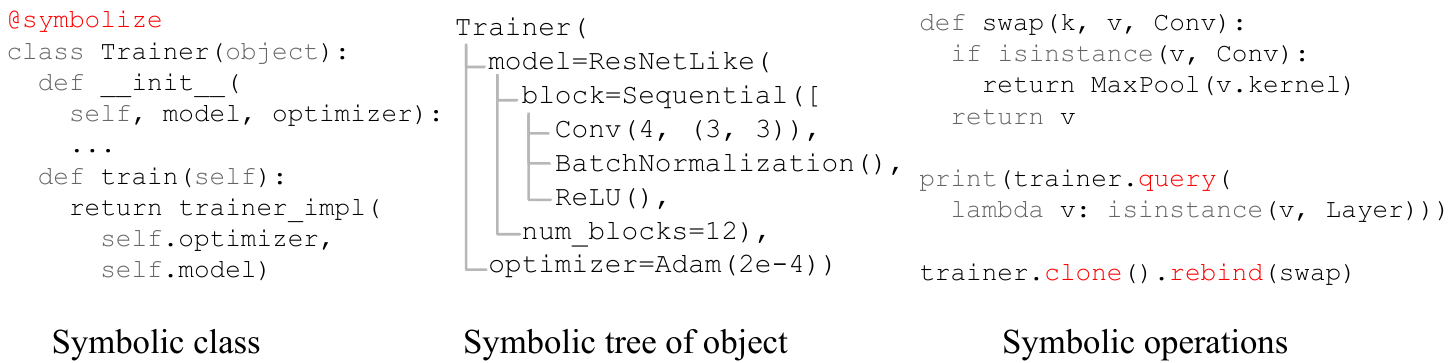}
\end{center}
\vspace{-5pt}
\caption{
\footnotesize{
A regular Python class made symbolically programmable via the \code{symbolize} decorator (left), whose object is a symbolic tree (middle), in which all nodes can be symbolically operated (right). For example, we can (i) retrieve all the \code{Layer} objects in the tree via \code{query}, (ii) \code{clone} the object and (iii) modify the copy by swapping all \code{Conv} layers with \code{MaxPool} layers of the same kernel size using \code{rebind}. \looseness=-1
}}
\label{fig:pyglove_program}
\vspace{-15pt}
\end{figure}

\subsection{Symbolize a Python program}\label{subsec:pyglove_program}

In PyGlove, preexisting Python programs can be made symbolically programmable with a \emph{symbolize} decorator. Besides classes, functions can be symbolized too, as discussed in Appendix B.1.2. To facilitate manipulation, PyGlove provides a wide range of symbolic operations. Among them, \emph{query}, \emph{clone} and \emph{rebind} are of special importance as they are  foundational to other symbolic operations. Examples of these operations can be found in Appendix B.2. \Figref{fig:pyglove_program} shows (1) a symbolic Python class, (2) an instance of the class as a symbolic tree, and (3) key symbolic operations which are applicable to a symbolic object. To convey the amount of work required to drop PyGlove into real-life projects,  we show the number of lines of code in making a PyTorch~\cite{paszke2019pytorch} and a TensorFlow~\cite{abadi2016tensorflow} projects searchable in \Tabref{table:drop_in_cost}.

\vspace{-5pt}
\subsection{From a symbolic program to a search space}
\label{subsec:pyglove_ssd}

\begin{wrapfigure}{r}{0.37\textwidth}
\vspace{-35pt}
\begin{center}
\includegraphics[width=\linewidth]{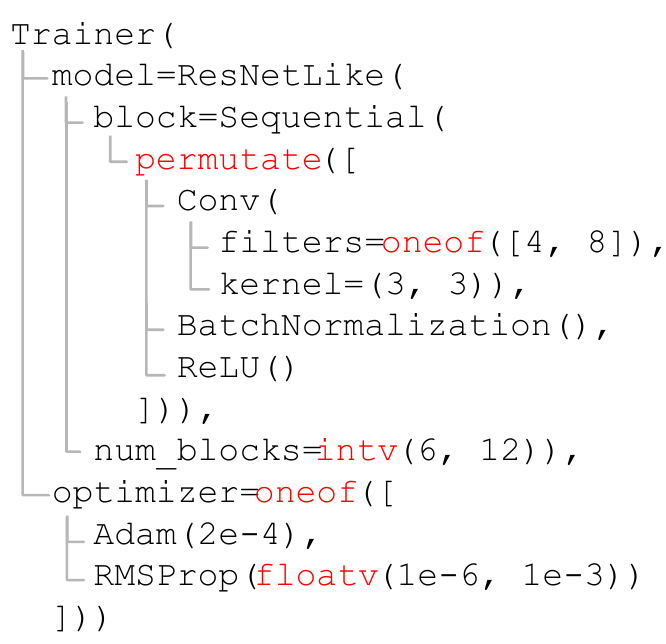}
\end{center}
\vspace{-5pt}
\caption{
\footnotesize{
The child program from \Figref{fig:pyglove_program}-2 is turned into a search space.\looseness=-1}}
\label{fig:pyglove_ssd}
\vspace{6pt}
\includegraphics[width=\linewidth]{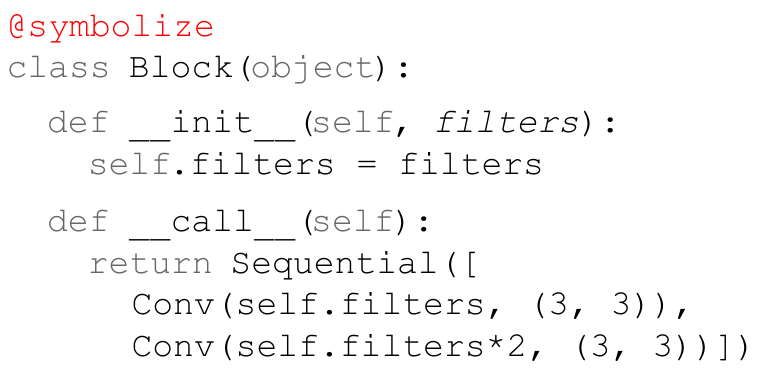}
\caption{\footnotesize{Expressing dependent hyper-parameters by introducing a higher-order symbolic \code{Block} class.\looseness=-1}}
\label{fig:pyglove_dependent_ssd}
\vspace{-20pt}
\end{wrapfigure}

With a child program being a symbolic tree, \emph{any node} in the tree can be replaced with a to-be-determined specification, which we call \emph{hyper value} (in correspondence to \emph{hyperify}, a verb introduced in Section~\ref{subsec:symbolic_process} in making search spaces). A search space is naturally represented as a symbolic tree with hyper values. In PyGlove, there are three classes of hyper values: 1) a continuous value declared by \code{floatv}; 2) a discrete value declared by \code{intv}; and 3) a categorical value declared by \code{oneof}, \code{manyof} or \code{permutate}.
\Tabref{table:hyper_values} summarizes different hyper value classes with their semantics. \Figref{fig:pyglove_ssd} shows a search space that jointly optimizes a model and an optimizer. The model space is a number of blocks whose structure is a sequence of permutation from [\code{Conv}, \code{BatchNormalization}, \code{ReLU}] with searchable filter size.



\emph{Dependent hyper-parameters} can be achieved by using higher-order symbolic objects. For example, if we want to search for the filters of a \code{Conv}, which follows another \code{Conv} whose filters are twice the input filters, we can create a symbolic \code{Block} class, which takes only one filter size -- the output filters of the first \code{Conv} -- as its hyper-parameters. When it's called, it returns a sequence of 2 \code{Conv} units based on its filters, as shown in \Figref{fig:pyglove_dependent_ssd}. The filters of the block can be a hyper value at construction time, appearing as a node in the symbolic tree, but will be materialized when it's called.

\vspace{-5pt}
\subsection{Search algorithms}
\label{subsec:pyglove_algorithm}

\vspace{-2pt}

Without interacting with the child program and the search space directly, the search algorithm in PyGlove repeatedly 1) proposes an abstract child program based on the abstract search space and 2) receives measured qualities for the abstract child program to improve future proposals. PyGlove implements many search algorithms, including Random Search, PPO and Regularized Evolution.

\vspace{-10pt}
\begin{table}[h!]
\centering
\footnotesize{
\caption{Hyper value classes and their semantics.}
\vspace{5pt}
\begin{tabular}{c c c}
 Strategy & Hyper-parameter annotation & Search space semantics \\
 \toprule
 \small{Continuous} & \texttt{floatv}(min, max) & A float value from $\mathbb{R}^{[min, max]}$ \\
 \midrule
 \small{Discrete} & \texttt{intv}(min, max) &An int value from $\mathbb{Z}^{[min, max]}$ \\
 \midrule
 \small{Categorical} & \texttt{oneof}(candidates) & Choose 1 out of N candidates\\
 & \texttt{manyof}(K, candidates, $\theta$) & \makecell{Choose K out of N candidates \\ with optional constraints $\theta$ on the \\ uniqueness and order of chosen candidates}\\
  & \texttt{permutate}(candidates)& 
  \makecell{A special case of \code{manyof} which \\ searches for a permutation of all candidates}\\
  \midrule
 \small{Hierarchical} & \makecell{(when a categorical hyper value \\ contains child hyper values)} & Conditional search space\\
 \bottomrule
\end{tabular}
\vspace{-10pt}
\label{table:hyper_values}
}
\end{table}

\subsection{Expressing search flows}\label{subsec:pyglove_search}

\begin{figure}
\centering
\begin{minipage}[b]{0.5\linewidth}
\centering
\includegraphics[width=0.9\linewidth]{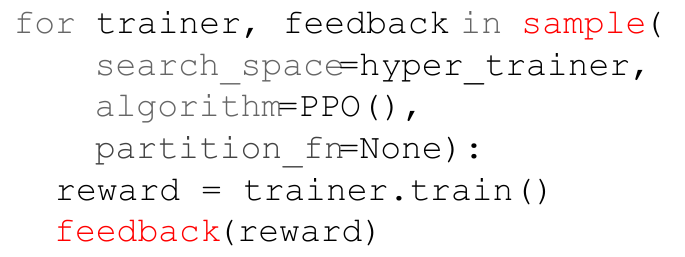}
\vspace{10pt}
\end{minipage}
\quad
\begin{minipage}[b]{0.35\linewidth}
\footnotesize{
\begin{tabular}{c l}
    Search type & for-loop pattern\\
    \toprule
    Joint & $\textcolor{blue}{\texttt{for}} (x, f_x): ...$\\
    \midrule
    \makecell{Separate}  & $\textcolor{blue}{\texttt{for}} (x_1, f_{x1}):...$\\
    & $\textcolor{blue}{\texttt{for}} (x_2, f_{x2}):...$\\
    \midrule
    \makecell{Factorized}       & $\textcolor{blue}{\texttt{for}} (x_1, f_{x1}):$\\
     & $\hspace{10pt}\textcolor{blue}{\texttt{for}} (x_2, f_{x2}):...$\\
    \bottomrule
\end{tabular}}
\label{table:search_flow}
\end{minipage}
\caption{\footnotesize{
PyGlove expresses search as a for-loop (left). Complex search flows can be expressed as compositions of for-loops (right).
}}
\label{fig:pyglove_search_flow}
\vspace{-17pt}
\end{figure}

With a search space, a search algorithm, and an optional search space partition function, a search flow can be expressed as a for-loop, illustrated in \Figref{fig:pyglove_search_flow}-left. Search space partitioning enables various ways in optimizing the divided sub-spaces, resulting in three basic search types: 1) optimize the sub-spaces \emph{jointly}; 2) optimize the sub-spaces \emph{separately}; or 3) \emph{factorize} the optimization. \Figref{fig:pyglove_search_flow}-right maps the three search types into different compositions of for-loop.

Let's take the search space defined in \Figref{fig:pyglove_ssd} as an example, which has a hyper-parameter sub-space (the hyper \code{optimizer}) and an architectural sub-space (the hyper \code{model}). Towards the two sub-spaces,
we can 1) jointly optimize them without specifying a partition function, as is shown in \Figref{fig:pyglove_search_flow}-left; 2) separately optimize them, by searching the hyper \code{optimizer} first with a fixed \code{model}, then use the best optimizer found to optimize the hyper \code{model}; or 3) factorize the optimization, by searching the hyper \code{optimizer} with a partition function in the outer loop. Each example in the loop is a trainer with a fixed \code{optimizer} and a hyper \code{model}; the latter will be optimized in the inner loop. The combination of these basic patterns can express very complex search flows, which will be further studied through our NAS-Bench-101 experiments discussed in Section~\ref{subsec:nasbench101}.


\vspace{-5pt}
\subsection{Switching between search spaces} \label{subsec:pyglove_modify_ssd}

\begin{wrapfigure}{r}{0.38\textwidth}
\vspace{-30pt}
\begin{center}
\includegraphics[width=\linewidth]{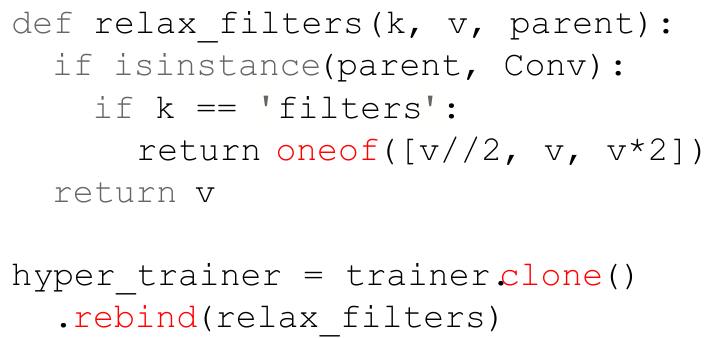}
\end{center}
\vspace{-5pt}
\caption{
\footnotesize{Manipulating the model in a trainer into a search space by relaxing the fixed filters of the \code{Conv} as a set of options.\looseness=-1
}}
\label{fig:pyglove_modify_search_space}
\vspace{-15pt}
\end{wrapfigure}

Making changes to the search space is a daily routine for AutoML practitioners, who may move from one search space to another, or to combine orthogonal search spaces into more complex ones.
For example, we may start by searching for different operations at each layer, then try the idea of searching for different output filters (\Figref{fig:pyglove_modify_search_space}), and eventually end up with searching for both. We showcase such search space exploration in \Secref{subsec:explore_search_space_and_algorithm}.

\vspace{-5pt}
\subsection{Switching between search algorithms}\label{subsec:pyglove_modify_algo}

The search algorithm is another dimension to experiment with. We can easily switch between search algorithms by passing a different algorithm to the \code{sample} function shown in Figure~\ref{fig:pyglove_search_flow}-1. When applying efficient NAS algorithms, the \code{hyper\_trainer} will be rewritten into a trainer that samples and trains the super-network transformed from the architectural search space.

\vspace{-5pt}
\section{Case Study}\label{sec:results}

\begin{wrapfigure}{r}{0.36\textwidth}
\vspace{-42pt}
\begin{center}
\includegraphics[width=\linewidth]{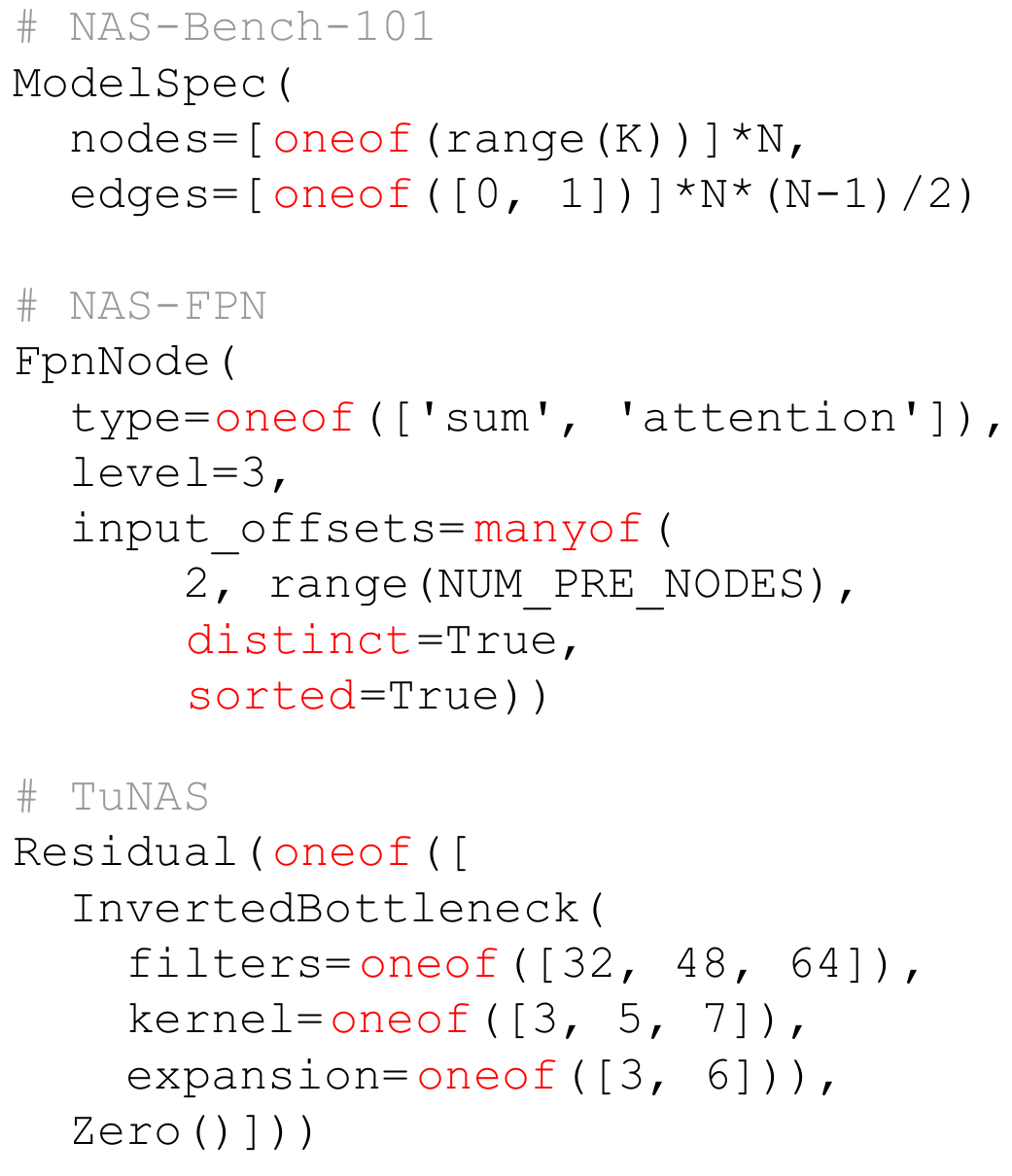}
\end{center}
\vspace{-4pt}
\caption{\footnotesize{Partial search space definition for {\nasone} (top), NAS-FPN (middle) and TuNAS (bottom).\looseness=-1}}
\label{fig:case_study_ssd}
\vspace{-50pt}
\end{wrapfigure}

In this section, we demonstrate that with PyGlove how users can define complex search spaces, explore new search spaces, search algorithms, and search flows with simplicity.

\subsection{Expressing complex search spaces} 

The composition of hyper values can represent complex search spaces. We have reproduced popular NAS papers, including {\nasone}~\cite{ying2019bench}, MNASNet~\cite{tan2019mnasnet}, NAS-FPN~\cite{nasfpn}, ProxylessNAS~\cite{cai2018proxylessnas}, TuNAS~\cite{bender2020tunas}, and NATS-Bench~\cite{dong2020nats}. 
Here we use the search spaces from {\nasone}, NAS-FPN, and TuNAS to demonstrate the expressiveness of PyGlove.

In the {\nasone} search space (Figure \ref{fig:case_study_ssd}-top), there are $N$ different positions in the network and ${N \choose 2} = \frac{N (N-1)}{2}$ edge positions that can be independently turned on or off. Each node independently selects one of $K$ possible operations.

The NAS-FPN search space is a repeated FPN cell, each of whose nodes (Figure \ref{fig:case_study_ssd}-middle) aggregates two outputs of previous nodes. The aggregation is either sum or global attention. We use \code{manyof} with the constraints \emph{distinct} and \emph{sorted} to select input nodes without duplication.

The TuNAS search space is a stack of blocks, each containing a number of residual layers (Figure \ref{fig:case_study_ssd}-bottom) of inverted bottleneck units, whose filter size, kernel size and expansion factor will be tuned. To search the number of layers in a block, we put \code{Zeros} as a candidate in the \code{Residual} layer so the residual layer may downgrade into an identity mapping.


\vspace{-5pt}
\subsection{Exploring search spaces and search algorithms}
\label{subsec:explore_search_space_and_algorithm}

We use MobileNetV2~\cite{sandler2018mobilenetv2} as an example to demonstrate how to explore new search spaces and search algorithms. For a fair comparison, we first retrain the MobileNetV2 model on ImageNet to obtain a baseline. 
With our training setup, it achieves a validation accuracy of 73.1\% (\Tabref{table:search_result}, row 1) compared with 72.0\% in the original MobileNetV2 paper. Details about our experiment setup, search space definitions, and the code for creating search spaces can be found in Appendix C.1.

\textbf{Search space exploration:~}
Similar to previous AutoML works~\cite{cai2018proxylessnas,tan2019mnasnet}, we explore 3 search spaces derived from MobileNetV2 that tune the hyper-parameters of the inverted bottleneck units~\cite{sandler2018mobilenetv2}:
(1) Search space $\sspace_1$ tunes the kernel size and expansion ratio.
(2) Search space $\sspace_2$ tunes the output filters
(3) Search space $\sspace_3$ combines $\sspace_1$ and $\sspace_2$ to tune the kernel size, expansion ratio and output filters.

From \Tabref{table:search_result}, we can see that with {\NAME} we were able to convert MobileNetV2 into $\sspace_1$ with \emph{23 lines} of code (row 2) and $\sspace_2$ with \emph{10 lines of code} (row 5). From $\sspace_1$ and $\sspace_2$, we obtain $\sspace_3$ in just \emph{a single line of code} (row 6) using \code{rebind} with chaining the transform functions from $\sspace_1$ and $\sspace_2$.

\vspace{-2pt}
\textbf{Search algorithm exploration:~}
On the search algorithm dimension, we start by exploring different search algorithms on $\sspace_1$ using black-box search algorithms (Random Search~\cite{bergstra2012random}, Bayesian~\cite{golovin2017google}) and then efficient NAS (TuNAS~\cite{bender2020tunas}). To make model sizes comparable, we constrain the search to 300M multiply-adds\footnote{For RS and Bayesian, we use rejection sampling to ensure sampled architectures have around 300M MAdds.} using TuNAS's absolute reward function \cite{bender2020tunas}. To switch between these algorithms, we only had to \emph{change 1 line of code}.

\vspace{-13pt}

\begin{table}[h!]
\small
\centering
\caption{\footnotesize{
Programming cost of switching between \textit{three} search spaces and \textit{three} AutoML algorithms based on PyGlove. Lines of code in red is the cost in creating new search spaces, while the lines of code in black is the cost for switching algorithms. The unit cost for search and training is defined as the TPU hours to train a MobileNetV2 model on ImageNet for 360 epochs. The test accuracies and MAdds are based on 3 runs.\looseness=-1
}}
\vspace{2pt}
\setlength{\tabcolsep}{5.4pt}
\begin{tabular}{c c c c c c c c c}
\# & Search space & \makecell{Search \\ algorithm}  & \makecell{\textbf{Lines}\\\textbf{of codes}} & \makecell{Search\\cost} & \makecell{Train\\cost}& \makecell{Test \\ accuracy} & \makecell{\# of\\MAdds}\\
\toprule
1 & ($static$) & N/A & N/A & N/A & 1 & 73.1 $\pm$ 0.1 & 300M \\
\midrule
2 & ($static$) $\rightarrow$ $\sspace_1$  & RS  & \textcolor{red}{\textbf{+23}}   & 25 & 1 &  73.7 $\pm$ 0.3 ($\uparrow$ 0.6) & 300 $\pm$ 3 M\\
3 & $\sspace_1$  & RS $\rightarrow$  Bayesian  & \textcolor{black}{\textbf{+1}}  & 25 & 1 &  73.9 $\pm$ 0.3 ($\uparrow$ 0.8) & 301 $\pm$ 5 M & \\
4 & $\sspace_1$  &           Bayesian $\rightarrow$ TuNAS  & \textcolor{black}{\textbf{+1}}  & 1 & 1 &  74.2 $\pm$ 0.1 ($\uparrow$ 1.1) & 301 $\pm$ 5 M&  \\
\midrule
5 & $(static) \rightarrow \sspace_2$     & TuNAS     & \textcolor{red}{\textbf{+10}} & 1 & 1 &  73.3 $\pm$ 0.1 ($\uparrow$ 0.2) & 302 $\pm$ 7M &  \\
\midrule
6 & $\sspace_1,\sspace_2 \rightarrow \sspace_3$ & TuNAS   & \textcolor{red}{\textbf{+1}}& 2 & 1 & 73.8 $\pm$ 0.1 ($\uparrow$ 0.7) & 302 $\pm$ 6M \\
\bottomrule
\end{tabular}
\label{table:search_result}
\end{table}

\vspace{-15pt}
\subsection{Exploring complex search flows on {\nasone}}
\label{subsec:nasbench101}

\begin{wrapfigure}{r}{0.38\textwidth}
\vspace{-40pt}
\begin{center}
\includegraphics[width=\linewidth]{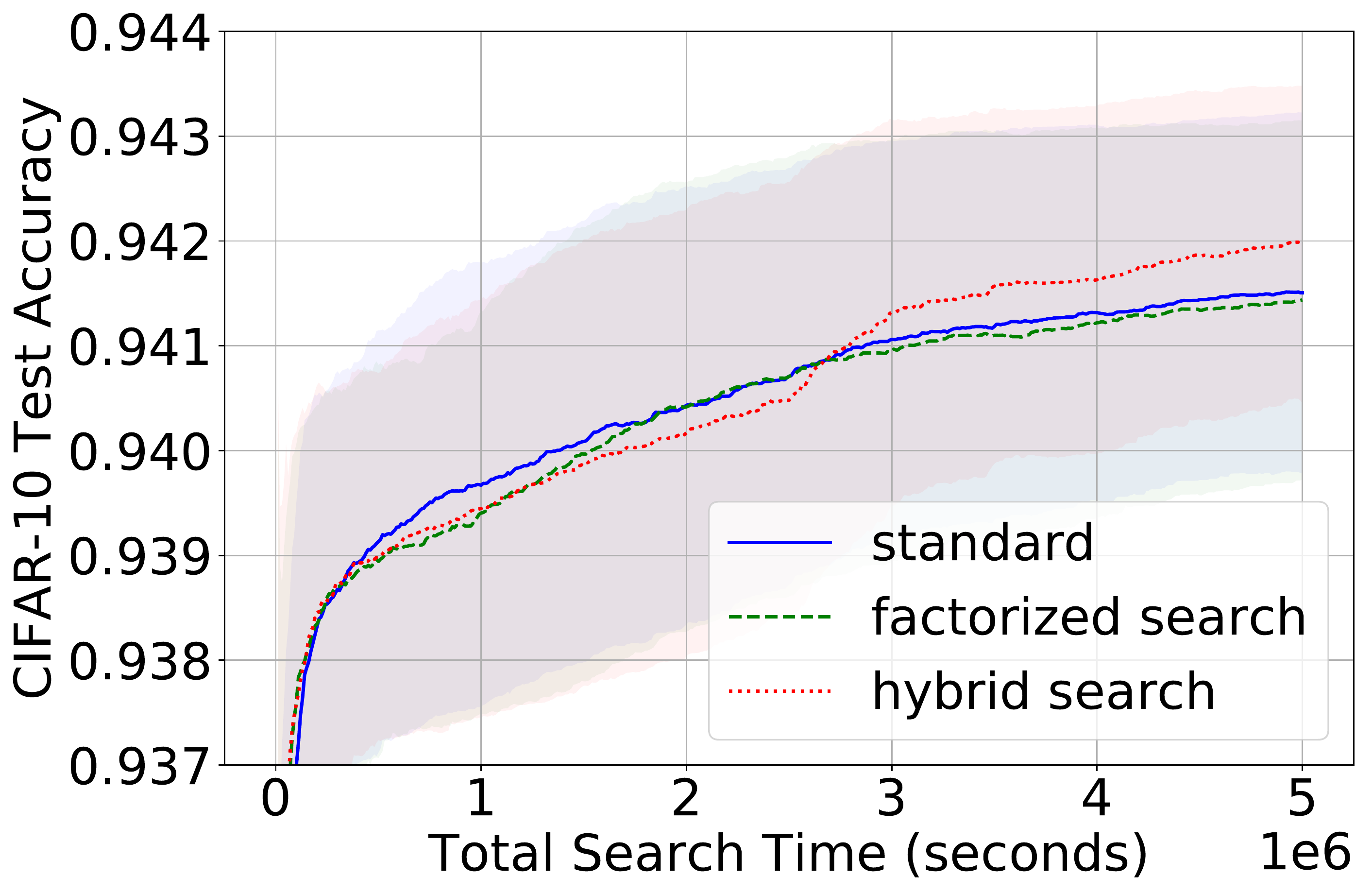}
\end{center}
\vspace{-2mm}
\caption{
\footnotesize{
\hspace{-5pt} Mean and standard deviation of search performances with different search flows on {\nasone} (500 runs), using Regularized Evolution~\cite{real2019regularized} .\looseness=-1
}}
\vspace{-17pt}
\label{fig:nas-bench-101}
\end{wrapfigure}
PyGlove can greatly reduce the engineering cost when exploring complex search flows. In this section, we explore various ways to optimize the {\nasone} search space. {\nasone} is a NAS benchmark where the goal is to find high-performing image classifiers in a search space of neural network architectures. This search space requires optimizing both the types of neural network layers used in the model (e.g., 3x3 Conv) and how the layers are connected.

We experiment with \emph{three} search flows in this exploration: 1) we reproduce the original paper to establish a baseline, which uses the search space defined in  \Figref{fig:case_study_ssd}-top to \emph{jointly} optimize the nodes and edges. 2) we try a \emph{factorized} search, which optimizes the nodes in the outer loop and the edges in the inner loop -- the reward for a node setting is computed as the average of top 5 rewards from the architectures sampled in the inner loop. While its performance is not as good as the baseline under the same search budget, we suspect that under each fixed node setting, the edge space is not explored enough. 3) To alleviate this problem, we come out a \emph{hybrid} solution, which uses the first half of the budget to optimize the nodes as in search flow 2, while using the other half to optimize the edges, based on the best node setting found in the first phase. Interestingly, the search trajectory crosses over the baseline in the second phase, ended with a noticeable margin (\Figref{fig:nas-bench-101}). We used  Regularized Evolution~\cite{real2019regularized} for all these searches, each with 500 runs. It takes only \emph{15 lines of code} to implement the factorized search and \emph{26 lines of code} to implement the hybrid search. Source codes are included in Appendix C.2.

\vspace{-5pt}
\section{Related Work}

\paragraph{Software frameworks} \hspace{-7pt} have greatly influenced and fueled the advancement of machine learning. The need for computing gradients has made auto-gradient based frameworks~\cite{bergstra2010theano, abadi2016tensorflow,paszke2019pytorch,jia2014caffe, tokui2018chainer,frostig2018compiling} flourish. To support modular machine learning programs with the flexibility to modify them,  frameworks were introduced with an emphasis on hyper-parameter management~\cite{shen2019lingvo, gincfg}. The sensitivity of machine learning to hyper-parameters and model architecture has led to the advent of AutoML libraries~\cite{NIPS2015_5872,kotthoff2017auto,feurer2019auto, golovin2017google,NNI,agtabular, negrinho2019towards,keras_tuner,jin2019auto}. Some (e.g.,~\cite{NIPS2015_5872,kotthoff2017auto, feurer2019auto}) formulate AutoML as a problem of jointly optimizing architectures and hyper-parameters. Others (e.g.,~\cite{golovin2017google,NNI,agtabular}) focus on providing interfaces for black-box optimization.
In particular, Google's Vizier library~\cite{golovin2017google} provides tools for optimizing a user-specified search space using black-box algorithms~\cite{bergstra2012random,jaderberg2017population}, but makes the end user responsible for translating a point in the search space into a user program. DeepArchitect~\cite{negrinho2019towards} proposes a language to create a search space as a program that connects user components. Keras-tuner~\cite{keras_tuner} employs a different way to annotate a model into a search space, though this annotation is limited to a list of supported components. Optuna~\cite{optuna_2019} embraces eager evaluation of tunable parameters, making it easy to declare a search space on the go (Appendix B.4).  Meanwhile, efficient NAS algorithms~\cite{pham2018efficient, liu2018darts,cai2018proxylessnas} brought new challenges to AutoML frameworks, which require coupling between the controller and child program. AutoGluon~\cite{agtabular} and NNI~\cite{NNI} partially solve this problem by building predefined modules that work in both general search mode and weight-sharing mode, however, supporting different efficient NAS algorithms are still non-trivial. Among the existing AutoML systems we are aware of, complex search flows are less explored. Compared to existing systems, PyGlove employs a mutable programming model to solve these problems, making AutoML easily accessible to preexisting ML programs. It also accommodates the dynamic interactions among the child programs, search spaces, search algorithms, and search flows to provide the flexibility needed for future AutoML research.

\vspace{-7pt}
\paragraph{Symbolic programming}

\hspace{-7pt}, where a program manipulates symbolic representations, has a long history dating back to LISP~\cite{mccarthy1960recursive}. The symbolic representation can be programs as in meta-programming, rules as in logic programming~\cite{colmerauer1996birth} and math expressions as in symbolic computation~\cite{Mathematica,buchberger1985symbolic}. In this work, we introduce the symbolic programming paradigm to AutoML by manipulating a symbolic tree-based representation that encodes the key elements of a machine learning program. Such program manipulation is also reminiscent of program synthesis~\cite{abelson1996structure, gulwani2011synthesis, gulwani2017program}, which searches for programs to solve different tasks like string and number manipulation~\cite{Polozov2015FlashMetaAF,Parisotto2017NeuroSymbolicPS,Devlin2017RobustFillNP,balog2017deepcoder}, question answering~\cite{Liang2018MemoryAP,Neelakantan2016NeuralPI}, and learning tasks~\cite{real2020automl,valkov2018houdini}. Our method also shares similarities with prior works in non-deterministic programming~\cite{Andre2002StateAI,sondergaard1992non,solar2009sketching}, which define non-deterministic operators like \emph{choice} in the programming environment that can be connected to optimization algorithms. Last but not least, our work echos the idea of building robust software systems that can cope with unanticipated requirements via advanced symbolic programming~\cite{Sussman2007BuildingRS}.

\vspace{-5pt}
\section{Conclusion}\label{sec:conclusion}
\vspace{-3pt}

In this paper, we reformulate AutoML as an automated process of manipulating a ML program through symbolic programming. Under this formulation, the complex interactions between the child program, the search space, and the search algorithm are elegantly disentangled. Complex search flows can be expressed as compositions of for-loops, greatly simplifying the programming interface of AutoML without sacrificing flexibility. This is achieved by resolving the conflict between AutoML's intrinsic requirement in modifying programs and the immutable-program premise of existing software libraries. We then introduce PyGlove, a general-purpose symbolic programming library for Python which implements our method and is tested on real-world AutoML scenarios. With PyGlove, AutoML can be easily dropped into preexisting ML programs, with all program parts searchable, permitting rapid exploration of different dimensions of AutoML.
\section*{Broader Impact}

Symbolic programming/PyGlove makes AutoML more accessible to machine learning practitioners, which means manual trial-and-error of many categories can be replaced by machines. This can also greatly increase the productivity of AutoML research, at the cost of increasing demand for computation, and -- a result -- increasing $\text{CO}_2$ emissions.


We see a big potential in symbolic programming/PyGlove in making machine learning researchers more productive. On a new ground of mutable programs, experiments can be reproduced more easily, modified with lower cost, and shared like data. A large variety of experiments can co-exist in a shared code base that makes combining and comparing different techniques more convenient.

Symbolic programming/PyGlove makes it much easier to develop search-based programs which can be used in a broad spectrum of research and product areas. Some potential areas, such as medicine design, have a clear societal benefit, while others potential applications, such as video surveillance, could improve security while raising new privacy concerns.

\begin{ack}
We would like to thank Pieter-Jan Kindermans and David Dohan for their help in preparing the case study section of this paper; Jiquan Ngiam, Rishabh Singh for their feedback to the early versions of the paper; Ruoming Pang, Vijay Vasudevan, Da Huang, Ming Cheng, Yanping Huang, Jie Yang, Jinsong Mu for their feedback at early stage of PyGlove; Adams Yu, Daniel Park, Golnaz Ghiasi, Azade Nazi, Thang Luong, Barret Zoph, David So, Daniel De Freitas Adiwardana, Junyang Shen, Lav Rai, Guanhang Wu, Vishy Tirumalashetty, Pengchong Jin, Xianzhi Du, Yeqing Li, Xiaodan Song, Abhanshu Sharma, Cong Li, Mei Chen, Aleksandra Faust, Yingjie Miao, JD Co-Reyes, Kevin Wu, Yanqi Zhang, Berkin Akin, Amir Yazdanbakhsh, Shuyang Cheng, HyoukJoong Lee, Peisheng Li and Barbara Wang for being early adopters of PyGlove and their invaluable feedback.

Funding disclosure: This work was done as a part of the authors' full-time job in Google.
\end{ack}

\bibliographystyle{unsrt}{\small
\bibliography{abrv,ms}
}

\newpage
\appendix
\begin{center}
\LARGE{\textbf{Appendix}}
\end{center}

The appendix provides a formal definition of symbolic programs in our method, including symbolic counterparts of different program constructs, supported operations, and the description of algorithms used in the materialization process. Appendix~\ref{sec:more_about_pyglove} gives a more detailed introduction to PyGlove -- our implementation of the method -- with an example of dropping neural architecture search (NAS) into an existing Tensorflow program (MNIST~\cite{abadi2016tensorflow}). Appendix~\ref{sec:more_about_experiments} provides additional information for experiments used in our case studies, including experiment setup, source code for creating search spaces and complex search flows.

\section{More on Symbolic Programming for AutoML}
\label{sec:more_about_our_method}
\subsection{Formal definition of a symbolic program}

Give a program construct type $t$, let the hyper-parameters (which defines the uniqueness of an instance of $t$) be noted as $P(t) = \langle p_0, ..., p_n\rangle$. The symbolic type of $t$ can then be defined as the output of the symbolization function $S$ applied on $t$, which returns a tuple of $t$'s type information and its hyper-parameter definitions:

\begin{equation}
s = S(t) = \langle t, P(t)\rangle
\label{eq:symbolize}
\end{equation}

A hyper-parameter $p_i$ of $s$ is either a primitive type or a symbolic type. Therefore an instance $x$ of $s$ -- a symbolic object -- is a tree node, whose sub-nodes are its hyper-parameters. For convenience, $x$ is called a symbolic $t$, e.g: symbolic \code{Dataset}, symbolic \code{Conv}, etc.  A symbolic program is a symbolic object that can be executed, for example, a symbolic \code{Trainer} that trains and evaluates a \code{ResNet} (as a sub-node) on ImageNet.

Two tree nodes are equal if and only if their type and hyper-parameters are equal. For example, consider a symbolic \code{Conv} class which takes \code{filters}, \code{kernel\_size} as its hyper-parameters. Two \code{Conv} instances are equal if and only if their \code{filters} and \code{kernel\_size} are equal.

We can clone a tree by copying its type information and hyper-parameters. Similarly, we can replace a hyper-parameter value with a new value, which is the foundation for symbolic manipulation. For example, a symbolic \code{Conv}'s \code{kernel\_size} can be changed from $(3, 3)$ to $(5, 5)$ by another program.

Symbolic constraints can be specified on the hyper-parameters. These constraints define the hyper-parameters' value types and ranges. When a value is assigned as a hyper-parameter of another symbolic object, it will be validated based on the symbolic constraint on that hyper-parameter. Since the sub-nodes of a symbolic object can be manipulated, the constraints are helpful in catching mistakes during symbolic manipulation. 

\subsection{Symbolic types}
The basic elements of a computer program are classes and functions, plus a few built-in data structure that works with the classes and functions for composition. To symbolize a computer program, we need to map these basic program constructs to their symbolic counterparts.
Based on Equation~\ref{eq:symbolize}, the symbolic type of $t$ is defined by $t$'s type information and hyper-parameters, illustrated in \Tabref{table:symbolic_hps}

\begin{table}[H]
\centering
\caption{Hyper-parameters of basic program constructs} 
\vspace{5pt}
\begin{tabular}{c c}
Program construct type & Hyper-parameters \\
\toprule
\texttt{class} & Constructor arguments. \\
\texttt{function} & Function arguments. \\
\texttt{list} & Indices in the list. \\
\texttt{dict} & Keys in the dict. \\
\bottomrule
\end{tabular}
\label{table:symbolic_hps}
\end{table}

Though a regular function takes arguments, the function itself doesn't hold its hyper-parameter values. Therefore, in order to manipulate the hyper-parameters of a function, a symbolic function -- functor -- behaves like an object: an function with bound arguments. As a result, a functor is no different from a class object with a call method, whose arguments could be bound either at construction time or call time. Therefore, a functor can be a node in the symbolic tree.

\subsection{Operations on symbolic types}

Symbolic objects can be manipulated via a set of operations. \Tabref{table:symbolic_ops} lists the basic operations applicable to all symbolic types. Particularly, \code{rebind} in the modification category is of special importance, as it's the foundation for implementing complex program transforms.

\begin{table}[H]
\vspace{-5pt}
\caption{Basic operations applicable to symbolic types.}
\vspace{5pt}
\centering
\begin{tabular}{c c c}
     Category & Operation & Description\\
     \toprule
     Modification & \texttt{rebind}($x, \code{dict}$) & Replace each node in $x$ whose path is a key in \code{dict} \\
     & \texttt{rebind}($x, \lambda$) & Recursively apply the function $\lambda$ to each node in $x$\\
     \midrule
     Inference & \texttt{isinstance}($x$, $t$) & Returns true if $x$ is an instance of $t$, false otherwise \\
     & \texttt{has}($x$, $p$) & Returns true if $p$ is a property of $x$, false otherwise \\
     & \texttt{equal}($x$, $x'$) & Returns true if $x$ equals $x'$, false otherwise \\
     \midrule
     Inquiry & \texttt{parent}($x$) & Returns the parent node of $x$\\
     & \texttt{path}($x$) & 
     Returns the path from the tree root to $x$ \\
     & \texttt{get}($x$, $l$) & Returns the sub-node of $x$ which has path $l$\\
     & \texttt{query}($x$, $\theta$)& \begin{tabular}{c}Returns a dict of $\langle \text{path}, \text{value} \rangle$ pairs which \\ contains all sub-nodes of $x$ satisfying predicate $\theta$\end{tabular} \\
     \midrule
     Replication & \texttt{clone}($x$) & Returns a symbolic copy of $x$\\
     \bottomrule
\end{tabular}
\label{table:symbolic_ops}
\end{table}
\vspace{-20pt}
\subsection{Materializing a child program from an abstract child program}
As we decouple the search algorithm from the search space and child program by introducing the \emph{abstract search space} and \emph{abstract child program}, we need to materialize the abstract child program into a concrete child program based on the search space. Algorithm~\ref{alg:decode} illustrates this process, which recursively merges the hyper values from the search space and the numeric choices from the abstract child program. For a continuous or discrete hyper value, the value of choice is the final value to be assigned to its target node in the tree, while for a categorical hyper value, the value of choice is the index of the selected candidate.

\begin{algorithm}[H]
    \caption{\texttt{materialize}}
    \label{alg:decode}
    \DontPrintSemicolon
    \SetArgSty{textup}
    \KwIn{$search\_space$, $abstract\_child\_program$}
    \KwOut{$child\_program$}
    \BlankLine
    \If{\texttt{isinstance}($search\_space$, \texttt{Choice})}{
        $xcs\gets ()$\;
        \ForAll{$c \in \texttt{child\_space}(search\_space)$}{
            $dc\gets \texttt{child\_value}(abstract\_child\_program, \texttt{path}(c))$\;
            $xc\gets \texttt{materialize}(c, dc)$\;
            $\texttt{append}(xcs, xc)$\;
        }
        $child\_program\gets xcs[\texttt{value\_of}(abstract\_child\_program)]$\;
    }
    \Else{
        $child\_program\gets \texttt{value\_of}(abstract\_child\_program)$\;
    }
    \Return $child\_program$
\end{algorithm}

\subsection{Sampling child programs from a search space}
Sampling a child program from a search space can be described as a process in which 1) the search algorithm proposes an abstract child program, and 2) the search space materializes the abstract child program into a concrete program. Before the process starts, an abstract search space will be obtained from the search space for setting up the search algorithm. This process is described by Algorithm~\ref{alg:sample_impl}.

\begin{algorithm}[H]
    \SetKw{Yield}{yield}
    \caption{\texttt{sample}}
    \DontPrintSemicolon
    \KwIn{$search\_space$, $search\_algorithm$}
    \KwOut{$\texttt{Iterator}(\langle child\_program, feedback\_for\_child\rangle)$}
    \BlankLine
     \texttt{setup}(search\_algorithm, \texttt{abstract\_search\_space}(search\_space))\;
    \While{$true$}{
        $abstract\_child\_program\gets \texttt{propose}(search\_algorithm)$\;
        $child\_program\gets \texttt{materialize}(search\_space, abstract\_child\_program)$\;
        $feedback\_for\_child \gets \texttt{partial\_bind}(\texttt{feedback}, search\_algorithm, abstract\_child\_program)$

        \Yield{$\langle child\_program, feedback\_for\_child\rangle$}
    }
    \label{alg:sample_impl}
\end{algorithm}

\clearpage
\section{More on PyGlove}\label{sec:more_about_pyglove}
In this section, we will map the concepts from our method into PyGlove programs, to illustrate how a regular Python program is made symbolic programmable, turned into a search space, and then optimized in a search flow. At the end of this section, we provide an example of enabling NAS for an existing Tensorflow-based MNIST program.

\subsection{Symbolize a child program}
\subsubsection{Symbolize classes}

A symbolic class can be converted from a regular Python class using the \code{@symbolize} decorator, or can be created on-the-fly without modifying the original class. The \code{symbolize} decorator creates a class on-the-fly by multi-inheriting the symbolic Object base class and the user class. The resulting class therefore possesses the capabilities of both parents. Figure~\ref{fig:symbolic_class_def} shows an code example of symbolizing existing/new classes.

\begin{center}
\begin{minipage}{0.7\textwidth}    
\captionsetup{type=figure}
\begin{lstlisting}[style = Python]
import pyglove as pg
import tensorflow as tf

# Symbolizing preexisting keras layers into symbolic 
# classes without modifying original classes.
Conv2D = pg.symbolize(tf.keras.layers.Conv2D)
Dense = pg.symbolize(tf.keras.layers.Dense)
Sequential = pg.symbolize(tf.keras.Sequential)

# Symbolizing a newly created class with constraints.
@pg.symbolize([
  ('learning_rate', pg.typing.Float(min_value=0)),
  ('steps', pg.typing.Int(min_value=1))
])
class CosineDecay(object):

  def __init__(self, learning_rate, steps):
    self.learning_rate = learning_rate
    self.steps = steps

  def __call__(self, current_step):
    return (tf.cos(np.pi * current_step / self.steps) 
            * self.learning_rate)

\end{lstlisting}
\captionof{figure}{Symbolizing existing classes and new classes.}
\label{fig:symbolic_class_def}
\end{minipage}
\end{center}

\paragraph{Using symbolic constraints}  Constraints which validate new values during object construction or upon modification can be optionally provided when using the \code{@symbolize} decorator. Symbolic constraints can greatly reduce human mistakes when a program is manipulated by other programs. It also make the program implementation more crisp: user can program against an argument as it claims to be without additional check.

\paragraph{Recomputing internal states} Symbolic objects may have internal states. The mutable programming model will only work when the internal states are consistent upon modification.
When one or more hyper-parameters are modified through \code{rebind}, the object's state will be reset, and the object's constructor will be invoked (again) on the same instance. Moreover, the change propagates back from the current node to the root of the symbolic tree, allowing all impacted nodes to recompute states upon modification.


\subsubsection{Symbolize functions}
\label{subsec:functors}
\paragraph{From function to functor}

Making functions symbolic programmable is trickier than for classes, for the following reasons: First: functions don't explicitly hold their parameters as member variables, although functions' bound arguments are analogous to member variables in classes. Second: functions don't have the concept of inheritance, which is necessary to get access to the capabilities provided by the symbolic Object base class. To address these two issues, we introduce the concept of \emph{functor}, which is a symbolic class with a \code{\_\_call\_\_} method; all the function arguments becoming the functor's hyper-parameters. Under the functor concept, we unify the representation and operations of classes and functions. Figure~\ref{fig:symbolize_function} shows that functions can be symbolized in the same way as we symbolize classes. Figure~\ref{fig:functor} shows how functors can be used with great flexibility in binding their hyper-parameters.

\vspace{-5pt}
\begin{center}
\begin{minipage}{0.7\textwidth}
\captionsetup{type=figure}
\begin{lstlisting}[style = Python]
@pg.symbolize
def random_augment(image, magnitude):
  return random_augment_impl(data, magnitude)

@pg.symbolize([
  ('model', pg.typing.Instance(Layer)),
  ('augment_policy', pg.typing.Callable(
     [pg.typing.Instance(tf.Tensor)], 
     returns=pg.typing.Instance(tf.Tensor))),
  ('learning_schedule', pg.typing.Callable([
     pg.typing.Instance(tf.Tensor)]))
])
def train_model(model, 
                augment_policy, 
                learning_schedule):
  return train_model_impl(
    model, augment_policy, learning_schedule)
\end{lstlisting}
\vspace{-5pt}
\captionof{figure}{Decorator \code{symbolize} converts functions into functors. Since properties for functors are automatically added from function signature, constraints are optional. Nevertheless, users are encouraged to add constraints for functor properties for safety and productivity.}
\label{fig:symbolize_function}
\end{minipage}
\end{center}
\vspace{-5pt}
\begin{center}
\begin{minipage}{0.7\textwidth}
\captionsetup{type=figure}
\begin{lstlisting}[style = Python]
model = Sequential(children=[
  Conv2D(filters=8, kernel_size=(3, 3)), 
  Dense(units=10)
])

# Partial parameter binding, in which `model` is missing.
trainer = train_model(
  augment_policy=random_augment(
    magnitude=8))

# Incremental parameter binding via assignment.
trainer.learning_schedule = CosineDecay(1e-5, 5000)

# Incremental parameter binding at call time.
accuracy1 = trainer(model=model)

# Call with overriding previously bound parameters.
accuracy2 = trainer(
  model=model,
  learning_schedule=CosineDecay(2e-4, 5000), 
  override_args=True)
\end{lstlisting}
\vspace{-10pt}
\captionof{figure}{Functors can be used as objects, with a rich set of argument binding features.}

\label{fig:functor}
\end{minipage}
\end{center}

\paragraph{Partial and incremental argument binding} Functor comes with a capability that allows arguments to be partially bound at construction time, incrementally bound via property assignment and at call time. We can even override a previously bound argument during the call to the functor.

\subsection{Operating symbolic values}

Symbolic values can be operated as if they were plain data, including inference, inquiry, modification and replication. Figure~\ref{fig:symbolic_ops_code} gives some examples to these operations.  

\begin{center}
\begin{minipage}{0.80\textwidth}
\captionsetup{type=figure}
\begin{lstlisting}[style = Python]
model = Sequential(children=[
  Conv2D(filters=8, kernel_size=(3, 3)), 
  Dense(units=10)
])

# Partial parameter binding, in which `model` is missing.
trainer = train_model(
  augment_policy=random_augment(
    magnitude=8))

# Inference.
assert isinstance(trainer, train_model)
assert isinstance(trainer.model, Layer)
assert trainer.model.children[1] == Dense(10)
assert trainer.model != Conv2D(16, (3, 3))

# Inquiry.
assert trainer.query('.*filters') == {
    'model.children[0].filters': 8
  }
assert trainer.query(where=(
  lambda v: isinstance(v, Dense))) == {
    'model.children[1]': Dense(units=10)
  }

# Modification.
assert trainer.rebind({
    'model.children[0].filters': 16,   
    'model.children[1]': insert(Dense(20)) 
  }).model == Sequential([
    Conv2D(16, (3, 3)), Dense(20), Dense(10)
  ])

def conv_to_dense(k, v):
  return Dense(v.filters) if isinstance(v.Conv2D) else v
assert trainer.rebind(conv_to_dense) == (
  Sequential([Dense(16), Dense(20), Dense(10)])

# Replication.
assert trainer.clone() == trainer
assert trainer.clone(deep=True) == trainer
trainer.save('trainer.json')
assert pg.load('trainer.json') == trainer
\end{lstlisting}
\vspace{-5pt}
\captionof{figure}{Example code for symbolic operations on inference, comparison, inquiry, modification, replication and serialization.}
\label{fig:symbolic_ops_code}
\end{minipage}
\end{center}

\subsection{Using PyGlove for search}
\subsubsection{Creating search spaces}
With the definition of functors \code{train\_model} and \code{random\_augment}, as well as the layer classes, we can create a search space by replacing concrete values with hyper values, illustrated in \Figref{fig:search_space_code}.

\begin{center}
\begin{minipage}{0.80\textwidth}
\captionsetup{type=figure}
\begin{lstlisting}[style = Python]
hyper_trainer = train_model(
  model=Sequential(
    pg.manyof(k=3, candidates=[
        Conv2D(filters=pg.oneof([8, 16]), 
               kernel_size=pg.oneof([(3, 3), (5, 5)])),
        Dense(units=pg.oneof([10, 20]))
    ], choices_distinct=False)),
  augment_policy=random_augment(
      magnitude=pg.oneof([3, 6, 9])),
  learning_schedule=CosineDecay(pg.floatv(1e-5, 1e-4), 5000))
\end{lstlisting}
\captionof{figure}{An example of conditional search space for jointly searching the model architecture, data augment policy, and learning rate.}
\label{fig:search_space_code}
\end{minipage}
\end{center}

\subsubsection{Search: putting things together}
\label{subsec:search_flow}
With \code{hyper\_trainer} as the search space, we can start a search by sampling concrete trainers from the search space with a search algorithm (e.g. \code{RegularizedEvolution}~\cite{real2019regularized}). The \code{trainer} is a concrete instance of \code{train\_model}, which can be invoked to return the validation accuracy on ImageNet. We use the validation accuracy as a reward to feedback to the search algorithm, illustrated in \Figref{fig:search_code}.

\begin{center}
\begin{minipage}{0.80\textwidth}
\captionsetup{type=figure}
\begin{lstlisting}[style = Python]
for trainer, feedback in pg.sample(
    hyper_trainer, pg.generators.RegularizedEvolution(),
    partition_fn=None):
  reward = trainer()
  feedback(reward)

\end{lstlisting}
\captionof{figure}{Creating a search flow from a search space and a search algorithm. We pass None to the search space partition function here as to optimize the whole search space.}
\label{fig:search_code}
\end{minipage}
\end{center}

\subsection{More on materialization of hyper values}
Materialization of hyper values can take place either eagerly or in a late-bound fashion. In the former case, the hyper value evaluates to a concrete value within its range upon creation, and register the search space into a global context for the first run, which can be picked up by the search algorithm later to propose values for future runs. This conditional evaluation makes it possible to support the define-by-run style search space definition advocated by Optuna~\cite{optuna_2019}. In the latter case, the search space will be inspected from the symbolic tree and the tree can be manipulated freely by the search algorithm before the program is executed. 

\begin{center}
\begin{minipage}{0.7\textwidth}    
\captionsetup{type=figure}
\begin{lstlisting}[style = Python]
def oneof(candidates, hints=None):
  """Oneof with optional eager execution."""
  choice = Choice(candidates, hints)
  if is_eager_mode():
     if is_apply_decisions():
        # Apply next decision from the global context.
        chosen_index = next_global_decision()
     else:
        # Collect the decision points when running
        # the program for the first time.
        add_global_decision_point(choice)
        chosen_index = 0
     choice = candidates[chosen_index]
  return choice
     
\end{lstlisting}
\captionof{figure}{Eagerly evaluation of hyper values.}
\label{fig:eager_evaluation}
\end{minipage}
\end{center}

The advantage of eager evaluation is that one can drop AutoML into a new ML program with minimal code changes. Users do not need to explicitly define the hyper-parameters to search. Instead, we can automatically identify them by executing the user's code before the start of the search. On the other hand, scattered searchable hyper-parameters makes it hard or error-prone to modify search space over many files, especially when we want to explore multiple search spaces. 

Meanwhile, conditional search spaces require special handling. Define-by-run semantics typically do not provide enough information for us to recognize hierarchical search spaces. For instance, it is difficult to distinguish between \code{oneof([oneof([1, 2]), 1])} and \code{oneof([1, 2]) + oneof([3, 4])}. In PyGlove, we solve this problem by using a lambda function with zero-argument which returns the candidate: \code{oneof([lambda:oneof([1, 2]), 1])}. In this case, the outer \code{oneof} will instantiate the inner \code{oneof}, making it possible to capture the hierarchy of the hyper value structure.

While eagerly evaluation of hyper values seems to override the mechanism of symbolic manipulation, it is not so for PyGlove: Under eager mode, PyGlove runs the user program once to collect the symbolic objects (like the hyper values) along the program flow, so we can access these objects, manipulate them and inject them back into the program for future runs. As a result, eagerly evaluation can be regarded as an interface for PyGlove to inspect and manipulate the implicit symbolic objects created during program execution.

\subsection{Example: Neural Architecture Search on MNIST}
This section shows a complete example of dropping PyGlove into an existing ML program as to enable NAS. Added code is highlighted with a light-yellow background.

\begin{center}
\begin{lstlisting}[style=Python, escapechar=^]
"""NAS on MNIST.

This is a basic working ML program which does NAS on MNIST.
The code is modified from the tf.keras tutorial here:
https://www.tensorflow.org/tutorials/keras/classification

(The tutorial uses Fashion-MNIST,
but we just use "regular" MNIST for these tutorials.)

"""

from absl import app
from absl import flags
import numpy as np
^\highlight{\textcolor{blue}{import} pyglove as pg}^
import tensorflow as tf

flags.DEFINE_integer(
    'max_trials', 10, 'Number of max trials for tuning.')

flags.DEFINE_integer(
    'num_epochs', 10, 'Number of epochs to train for each trail.')

FLAGS = flags.FLAGS

def download_and_prep_data():
  """Download dataset and scale to [0, 1].

  Returns:
    tr_x: Training data.
    tr_y: Training labels.
    te_x: Testing data.
    te_y: Testing labels.
  """
  mnist_dataset = tf.keras.datasets.mnist
  (tr_x, tr_y), (te_x, te_y) = mnist_dataset.load_data()
  tr_x = tr_x / 255.0
  te_x = te_x / 255.0
  return tr_x, tr_y, te_x, te_y

# Create symbolized Keras layers classes.}
^\highlight{Conv2D = pg.\textcolor{red}{symbolize}(tf.keras.layers.Conv2D)}^
^\highlight{Dense = pg.\textcolor{red}{symbolize}(tf.keras.layers.Dense)}^
^\highlight{Sequential = pg.\textcolor{red}{symbolize}(tf.keras.Sequential)}^


def model_builder():
  """Model search space."""
  ^\highlight{return Sequential(pg.\textcolor{red}{oneof}([}^
    # Model family 1: only dense layers.
    ^\highlight{[}^
       ^\highlight{tf.keras.layers.Flatten()}^, 
       ^\highlight{Dense(pg.\textcolor{red}{oneof}([64, 128]), pg.\textcolor{red}{oneof}(['relu', 'sigmoid']))}^
    ^\highlight{]}^,
    # Model family 2: conv net.
    ^\highlight{[}^
       ^\highlight{tf.keras.layers.Lambda(lambda x: tf.reshape(x, (-1, 28, 28, 1)))}^,
       ^\highlight{Conv2D(pg.\textcolor{red}{oneof}([64, 128]), pg.\textcolor{red}{oneof}([(3, 3), (5, 5)])}^, 
             ^\highlight{activation=pg.\textcolor{red}{oneof}(['relu', 'sigmoid']))}^,
       ^\highlight{tf.keras.layers.Flatten()}^
    ^\highlight{]]) + [
      tf.keras.layers.Dense(10, activation='softmax')
    ])}^


def train_and_eval(model, input_data, num_epochs=10):
  """Returns model accuracy after train and evaluation.
  
  Args:
    model: A Keras model.
    input_data: A tuple of (training features, training_labels,
      test features, test labels) as input data.
    num_epochs: Number of epochs to train model.

  Returns:
    Accuracy on test split.
  """
  tr_x, tr_y, te_x, te_y = input_data
  model.compile(optimizer='adam',
                loss='sparse_categorical_crossentropy',
                metrics=['accuracy'])

  model.fit(tr_x, tr_y, epochs=num_epochs)
  _, test_acc = model.evaluate(te_x, te_y, verbose=2)
  return test_acc


def search(max_trials, num_epochs):
  """Search MNIST model via PPO.
  
  Args:
    max_trials: Max trials to search.
    num_epochs: Number of epochs to train individual trial.
  """
  results = []
  input_data = download_and_prep_data()
  ^\highlight{\textcolor{blue}{for} i, (model, \textcolor{red}{feedback}) \textcolor{blue}{in} enumerate(pg.\textcolor{red}{sample}(}^
      ^\highlight{model\_builder(), pg.generators.PPO(), max\_trials)):}^
    ^\highlight{test\_acc = train\_and\_eval(model, input\_data, num\_epochs)}^
    ^\highlight{results.append((i, test\_acc))}^
    ^\highlight{\textcolor{red}{feedback}(test\_acc)}^

  # Print best results.
  ^\highlight{top\_results = sorted(results, key=lambda x: x[2], reverse=True)}^
  ^\highlight{\textcolor{blue}{for} i, (trial\_id, test\_acc) \textcolor{blue}{in} enumerate(top\_results[:10]):}^
    ^\highlight{print('\{0:2d\} - trial \{1:2d\} (\{2:.3f\})'.format(
        i + 1, trial\_id, test\_acc))}^


def main(argv):
  """Program entrypoint."""
  if len(argv) > 1:
    raise app.UsageError('Too many command-line arguments.')
  ^\highlight{search(FLAGS.max\_trials, FLAGS.num\_epochs)}^

if __name__ == '__main__':
  app.run(main)
\end{lstlisting}
\end{center}

\clearpage
\section{More on case studies}
\label{sec:more_about_experiments}
This section describes the experiment details for our case studies in the paper.

\subsection{Search spaces and search algorithms exploration}

\subsubsection{Experiment setup}
\begin{table}[H]
    \centering
    \caption{Hyper-parameters for training MobileNetV2 and searched models.}
    \vspace{10pt}
    \begin{tabular}{cc}
         Name & Value \\
         \toprule
         Image size & 224 * 224 \\
         Pre-processing & ResNet preprocessing\\
         \midrule
         Training epochs & 360  \\
         Batch size & 4096 \\
         Optimizer& RMSProp: momentum 0.9, decay 0.9, epsilon 1.0 \\
         Learning schedule& \makecell{Cosine decay with peak learning rate 2.64, \\with 5 epochs linear warmup at the beginning.} \\
         L2 & 2e-5 \\
         \midrule
         Dropout rate & 0.15 \\
         Batch normalization & momentum 0.99, epsilon 0.001 \\
         \bottomrule
    \end{tabular}
    \label{table:training_setup}
\end{table}
\vspace{-5pt}
\begin{table}[H]
    \centering
    \caption{Definition of search spaces for exploration.}
    \begin{tabular}{c c}
         Search space & Hyper-parameters \\
         \toprule
         $\sspace_1$& Search the kernel sizes with candidates [(3, 3), (5, 5), (7, 7)] \\&  and expansion ratios with candidates [3, 6] for the inverted bottleneck units \\& in MobileNetV2; can remove layers from each block. \\
         \midrule
         $\sspace_2$& Search output filters of MobileNetV2 \\ & with multipliers [0.5, 0.625, 0.75, 1.0, 1.25, 1.5, 2.0] \\
         \midrule
         $\sspace_3$& Combine $\sspace_1$ and $\sspace_2$\\
         \bottomrule
    \end{tabular}
    \label{table:search_space_def}
\end{table}

\vspace{-10pt}
\begin{table}[H]
    \centering
    \caption{Search algorithm setup. We uses TuNAS absolute reward function with exponent=-0.1.}
    \vspace{10pt}
    \begin{tabular}{c c}
        Search algorithm & Configuration \\
        \toprule
        Random Search~\cite{bergstra2012random} & 100 trials, each trial trains for 90 epochs, \\ &rejection threshold for MAdds: $\pm$ 6M \\
        \midrule
        Bayesian Optimization~\cite{golovin2017google} & 100 trials, each trial trains for 90 epochs\\
        &rejection threshold for MAdds: $\pm$ 30M \\ & (We use a larger rejection ratio for Bayesian \\ &to limit the rejection rate, since our infra \\ & will take the rewards from rejected trials) \\
        \midrule
        TuNAS~\cite{bender2020tunas} & \makecell{Search for 90 epochs, with RL learning rate set to 0 for first 1/4 of training. \\The search cost for $\sspace_1$ and $\sspace_2$ is about 4x of static model training, \\and the cost for $\sspace_3$ about 8x of static model training.} \\
        \bottomrule
    \end{tabular}
    \label{tab:my_label}
\end{table}

\clearpage
\subsubsection{Creating search spaces}
In the Result section, we demonstrated 3 search spaces created from  MobileNetV2~\cite{sandler2018mobilenetv2}. Figure~\ref{figure:relax_ops}-\ref{figure:relax_both} show the code for converting the static model into search spaces $\sspace_1$, $\sspace_2$ and $\sspace_3$.

\vspace{-5pt}
\begin{center}
\begin{minipage}{0.9\textwidth}
\captionsetup{type=figure}
\begin{lstlisting}[style = Python]
import pyglove as pg

# Get the first inverted bottleneck.
r = model.query(lambda x: isinstance(x, InvertedBottleneck))
r = next(iter(r.values()))

def hyper_inverted_bottleneck(
  kernel_size_list, expansion_ratio_list, add_zeros=False):
    return pg.oneof([
      r.clone().rebind(kernel_size=k, expansion_ratio=e)
      for k in kernel_size_list
      for e in expansion_ratio_list
    ] + ([Zeros()] if add_zeros else []))

def relax_ops(k, v, p):
  if not k or k.key != 'op':
    return v
  # Check if the layer of current operation is the
  # first layer of current block.
  if k.parent.key == 0:
    if k == 'blocks[0].layers[0].op':
      return hyper_inverted_bottleneck(
        [(3, 3), (5, 5), (7, 7)], [1])
    else:
      return hyper_inverted_bottleneck(
        [(3, 3), (5, 5), (7, 7)], [3, 6])
  else:
    return hyper_inverted_bottleneck(
      [(3, 3), (5, 5), (7, 7)], [3, 6], True)

mobile_s2 = mobilenet_v2.clone().rebind(relax_ops)
\end{lstlisting}
\captionof{figure}{Added code for converting MobileNetV2 into a search space ($\sspace_1$) that tunes the kernel size and expansion ratio in all inverted bottleneck units.}
\label{figure:relax_ops}
\end{minipage}
\end{center}
\begin{center}
\begin{minipage}{0.9\textwidth}
\captionsetup{type=figure}
\begin{lstlisting}[style = Python]
def relax_filters(k, v, p):
  if isinstance(p, InvertedBottleNeck) and k == 'filters':
    scaled_values = sorted(set([
      layers.scale_filters(v, x)()
      for x in [0.5, 0.625, 0.75, 1.0, 1.25, 1.5, 2.0]]))
    if len(scaled_values) == 1:
      return scaled_values[0]
    return pg.oneof(scaled_values)
  return v
mobile_s1 = mobilenet_v2.clone().rebind(relax_filters)
\end{lstlisting}
\captionof{figure}{Added code for converting MobileNetV2 into a search space ($\sspace_2$) that tunes the channel size in all inverted bottleneck units.}
\label{figure:relax_filters}
\end{minipage}
\end{center}
\begin{center}
\begin{minipage}{0.8\textwidth}
\captionsetup{type=figure}
\begin{lstlisting}[style = Python]
mobile_s3 = mobilenet_v2.clone().rebind([relax_filters,    
                                         relax_ops])
\end{lstlisting}
\captionof{figure}{Applying transform functions from $\sspace_1$ and $\sspace_2$ to create $\sspace_3$.}
\label{figure:relax_both}
\end{minipage}
\end{center}

\subsection{Search flow exploration}
In the case study, we explored 3 search flows for optimizing NAS-Bench-101. Here we include the code for the factorized and hybrid search since the standard search is already discussed in \Secref{subsec:search_flow}.

\subsubsection{Factorized search}
\label{subsec:factorized_search}

For the  \emph{factorized} search, we optimize the nodes in the outer loop and the edges in the inner loop. Each example in the outer loop is a search space of edges with a fixed node setting. Each example in the inner loop is a fixed model architecture. The reward for the outer loop is computed as the average of top 5 rewards from the inner loop.

\begin{figure}[H]
\footnotesize
\begin{lstlisting}
def factorized_search(search_space):
  # Optimize the ops in the outer loop.
  # Each example in the outer loop is an edge sub-space with fixed
  # ops. `partition_fn` is used to create a sub-space by selecting
  # op hyper values only.
  best_example, best_reward = None, None
  for edge_space, ops_feedback in pg.sample(
      search_space, RegularizedEvolution(), 
      trials=300, partition_fn=lambda v: v.hints == OP_HINT):
    # Optimize the edges in the inner loop.
    # Each reward computed in the inner loop 
    # is for an edge setting relative to 
    # the node setting decided in the outer loop.
    rewards = []
    for example, edges_feedback in pg.sample(
        edge_space, RegularizedEvolution(), trials=20):
      reward = nasbench.get_reward(example)
      edges_feedback(reward)
      rewards.append(reward)
      if best_reward is None or best_reward < reward:
        best_example, best_reward = example, reward
    ops_feedback(top5_average(rewards))
  return best_example
\end{lstlisting}
\captionof{figure}{A factorized search that optimizes the nodes in the outer loop and the edges in the inner loop.}
\label{fig:tri_optimization}
\end{figure}
\clearpage
\subsubsection{Hybrid search}

For the \emph{hybrid} search, we use the first half of the budget to optimize the nodes using the same search flow illustrated in \Secref{subsec:factorized_search} , then we use the other half of the budget to further optimize the edges with the best nodes found in the prior phase.

\begin{figure}[H]
\footnotesize
\begin{lstlisting}
def hybrid_search(search_space):
  # Phase 1: search for the best ops with sampled edges. 
  # Each example in the outer loop is an edge sub-space with fixed
  # ops. `partition_fn` is used to create a sub-space by selecting
  # op hyper values only.
  ops_attempts = []
  for edge_space, ops_feedback in pg.sample(
      search_space, RegularizedEvolution(), 
      trials=150, partition_fn=lambda v: v.hints == OP_HINT):
    rewards = []
    algo = RegularizedEvolution()
    for example, edges_feedback in pg.sample(edge_space, 
                                             algo, trials=20):
      reward = nasbench.get_reward(example)
      edges_feedback(reward)
      rewards.append(reward)
    ops_reward = top5_average(rewards)
    ops_attempts.append((edge_space, ops_reward, algo))
    ops_feedback(ops_reward)
  
  # Phase 2: Continue search the best edge sub-space 
  # with best ops found.
  edge_space, _, edge_algo = sorted(
      ops_attempts, key=lambda x: x[1], reverse=True)[0]

  best_example, best_reward = None, None
  for example, edges_feedback in pg.sample(edge_space, 
                                           edge_algo, 150 * 20):
    reward = nasbench.get_reward(example)
    edges_feedback(reward)
    if best_reward is None or best_reward < reward:
      best_example, best_example = example, reward
  return best_example
\end{lstlisting}

\captionof{figure}{A hybrid search that optimizes the nodes with a factorized search in the first phase, and optimize the edges based on the best nodes found in the second phase.}
\label{fig:hybrid_search}
\end{figure}

\end{document}